\documentclass[letterpaper]{article} 
\usepackage{aaai25}  
\usepackage{times}  
\usepackage{helvet}  
\usepackage{courier}  
\usepackage[hyphens]{url}  
\usepackage{graphicx} 
\usepackage{natbib}  
\usepackage{caption} 
\usepackage{algorithm}
\usepackage{algorithmic}

\usepackage{amsmath}
\usepackage{amssymb}
\usepackage{array,booktabs}
\usepackage{tabularx}

\usepackage{newfloat}
\usepackage{listings}
\DeclareCaptionStyle{ruled}{labelfont=normalfont,labelsep=colon,strut=off} 
\floatstyle{ruled}
\newfloat{listing}{tb}{lst}{}
\floatname{listing}{Listing}
\title{Collaborative Learning for 3D Hand-Object Reconstruction and Compositional Action Recognition from Egocentric RGB Videos Using Superquadrics}
\author{
    Tze Ho Elden Tse\textsuperscript{\rm 1}, Runyang Feng\textsuperscript{\rm 2}, Linfang Zheng\textsuperscript{\rm 1}, Jiho Park\textsuperscript{\rm 3}, Yixing Gao\textsuperscript{\rm 2}\thanks{Corresponding author.}, Jihie Kim\textsuperscript{\rm 3}, \\ Ale\u{s} Leonardis\textsuperscript{\rm 1}, Hyung Jin Chang\textsuperscript{\rm 1}
}
\affiliations{
    \textsuperscript{\rm 1}University of Birmingham\quad \textsuperscript{\rm 2}Jilin University\quad \textsuperscript{\rm 3}Dongguk University\\

}

\usepackage{bibentry}

\newcommand{\comment}[1]{}

\newcommand{\bx}[0]{\mathbf{x}}

\newcommand{\fig}[1]{Fig.~\ref{fig:#1}}

\newcommand{\loss}{\mathcal{L}}

\newcommand{\eq}[1]{Eq.~\eqref{eq:#1}}

\newcommand{\eg}{{\it e.g. }}
\newcommand{\ie}{{\it i.e. }}

\renewcommand{\paragraph}[1]{\vspace{0.5em}\noindent\textbf{#1}}

\begin{document}

\maketitle

\begin{abstract}
With the availability of egocentric 3D hand-object interaction datasets, there is increasing interest in developing unified models for hand-object pose estimation and action recognition. However, existing methods still struggle to recognise seen actions on unseen objects due to the limitations in representing object shape and movement using 3D bounding boxes. Additionally, the reliance on object templates at test time limits their generalisability to unseen objects. To address these challenges, we propose to leverage superquadrics as an alternative 3D object representation to bounding boxes and demonstrate their effectiveness on both template-free object reconstruction and action recognition tasks. Moreover, as we find that pure appearance-based methods can outperform the unified methods, the potential benefits from 3D geometric information remain unclear. Therefore, we study the compositionality of actions by considering a more challenging task where the training combinations of verbs and nouns do not overlap with the testing split. We extend H2O and FPHA datasets with compositional splits and design a novel collaborative learning framework that can explicitly reason about the geometric relations between hands and the manipulated object. Through extensive quantitative and qualitative evaluations, we demonstrate significant improvements over the state-of-the-arts in (compositional) action recognition.
\end{abstract}

\section{Introduction}
\begin{figure}
\centering
\includegraphics[width=0.9\linewidth]{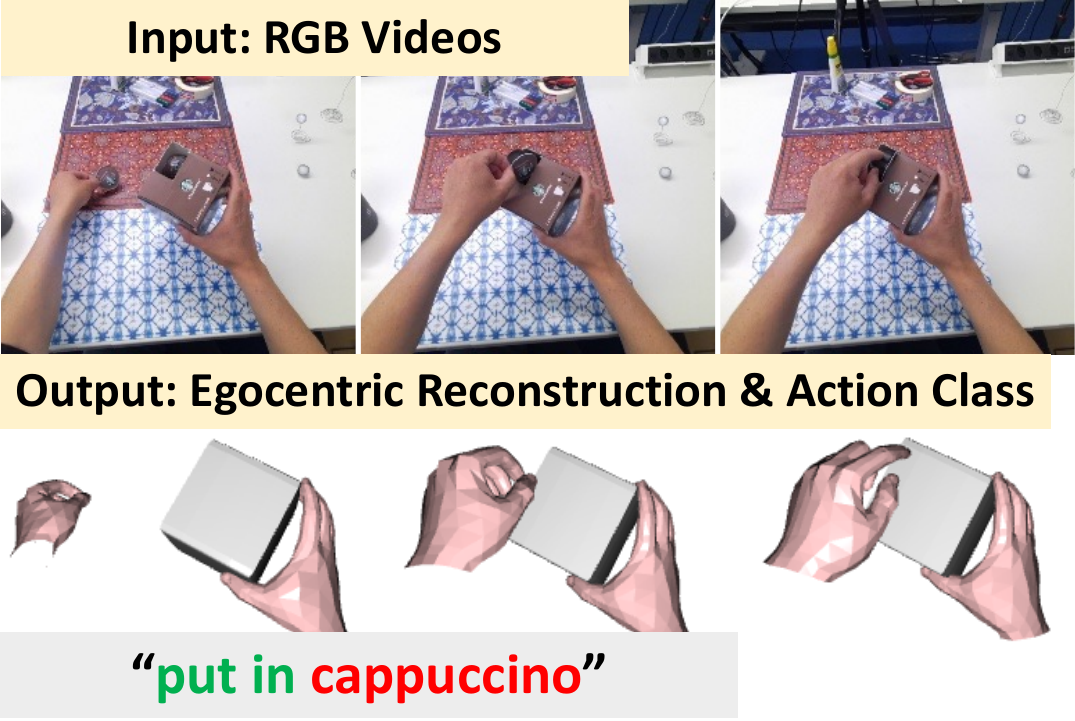}
\caption{Our method jointly reconstructs hand-object meshes without object instance-specific templates and recognises interaction from egocentric RGB video. We further consider a more challenging problem scenario, compositional action recognition, where combinations of verb (in green) and noun (in red) are unseen during training. Our model is designed for generalising action recognition by explicitly leveraging $3$D geometric information.}
\label{fig:teaser}
\end{figure}                                 

Understanding hand and object interaction is a fundamental problem in various downstream applications including augmented and virtual reality (AR/VR)~\cite{han2020megatrack,mueller2019real,han2022umetrack, Tse_2023_ICCV}. While a significant amount of research has focused on estimating the poses of hand and objects~\cite{tse2022collaborative,tse2022s, hampali2022keypoint, Yang_2022_CVPR,ye2022s,chen2022alignsdf,fan2023arctic,chen2023gsdf,ye2023affordance,feng2023diffpose,feng2023mutual}, jointly recognising hand-object interactions has received far less attention. In this paper, as illustrated in \fig{teaser}, we study jointly the problem of hand-object reconstruction and interaction recognition from egocentric view.

Recent egocentric hand-object interaction datasets~\cite{kwon2021h2o,garcia2018first} with $3$D annotations enable the development of a unified framework for estimating hand-object poses and interaction classes~\cite{tekin2019h+,yang2020collaborative,kwon2021h2o,wen2023hierarchical,cho2023transformer}. These methods couple $3$D geometric cues (\ie hand-object poses and/or contact maps) and appearance features to predict interaction class. Despite a unified understanding of the hand and manipulated object dynamics being crucial for recognising egocentric interactions, we find that pure appearance-based methods (\ie MViT~\cite{fan2021multiscale,li2022mvitv2}) can achieve comparable performance to the state-of-the-arts (as shown in Table~\ref{tbl:ch5:action}). This raises immediate questions on when or how $3$D geometric features can benefit interaction recognition.

In contrast, while deep architectures trained on large-scale datasets~\cite{sigurdsson2016hollywood,kay2017kinetics,karpathy2014large} exhibit strong distribution learning capabilities, mainstream action recognition models~\cite{simonyan2014two,carreira2017quo,feichtenhofer2019slowfast,wang2016temporal} primarily focus on frame appearance rather than temporal reasoning. Consequently, reversing the order of the video frame at test time will often produce the same classification result~\cite{materzynska2020something,zhou2018temporal}. In particular, classical activity recognition methods like the two-stream Convolutional Neural Network~\cite{simonyan2014two} and I$3$D~\cite{carreira2017quo} have demonstrated strong performance on various video datasets, including UCF$101$~\cite{soomro2012ucf101} and Sport$1$M~\cite{karpathy2014large}, with only still frames and optical flow. While appearance features can be highly predictive of the action class~\cite{santoro2017simple,battaglia2018relational}, it remains challenging for appearance-based deep networks to capture the \emph{compositionality} of action and objects without temporal transformations or geometric relations~\cite{materzynska2020something}.

To address the aforementioned problems, \citet{materzynska2020something} extends the Something-Something dataset~\cite{goyal2017something} and introduces the Something-Else task with a new compositional split. This presents a novel task known as compositional action recognition, in which methods are required to recognise an action with unseen objects. Under this problem setting, the combinations of actions and object instances do not overlap in the training and testing split. Therefore, models are encouraged to learn the compositionality of action \textit{verb} and \textit{noun}, and not overfit to the correlation between appearance features and action classes. Nonetheless, the current research in this task is primarily generic approaches using $2$D geometric cues such as $2$D instance bounding boxes. The potential benefits offered by $3$D geometric information remain an open problem.


Therefore, we take an alternative approach which exploits the compositionality of actions using $3$D geometric information. To achieve that, we first extend the two existing $3$D annotated egocentric hand-object datasets, H$2$O~\cite{kwon2021h2o} and FPHA~\cite{garcia2018first}, by introducing new compositional splits. Our experiments show that the existing state-of-the-art approaches~\cite{yang2020collaborative,wen2023hierarchical,fan2021multiscale,li2022mvitv2} still face significant challenges in recognising a seen action when facing new objects. This is because the current methods (either single or dual branches) are unable to tackle the problem of appearance bias in objects, as they have to take the combination of appearance and geometric information as a whole. In addition, these approaches focus on extracting features for the whole scene and do not explicitly recognise objects as individual entities. Hence, they cannot fully capture the compositionality of the action.

In this paper, we propose a collaborative learning framework that allows an action verb and object to interact and complement each other. The key motivation for this strategy is that the tasks of estimating hand-object poses and recognising interactions are naturally closely-correlated. Existing collaborative learning methods~\cite{yang2020collaborative,tse2022collaborative} in understanding hand-object interactions typically follow an iterative approach where the multiple target learning tasks (\ie hand-object pose estimation and action recognition) boost each other mutually and progressively. However, connecting branches iteratively can lead to highly unstable training~\cite{tse2022collaborative}. This is because gradients from one branch can propagate through the connections to affect the other branches which causes unstable gradients. We explicitly address this by a new transformer-based design to exploit the compositionality of actions and avoid branch stacking. In addition, we propose to use superquadrics~\cite{barr1981superquadrics} as the intermediate $3$D object representation. This is motivated by the fact that existing action recognition methods have limitations in accurately representing objects' shape and movement with only $2$D/$3$D bounding boxes. But at the same time, accurately reconstructing $3$D objects without object templates remains highly challenging, especially in scenarios involving unseen objects. Therefore, superquadrics offer a compact representation with their ability to represent a wide range of shapes with few parameters. In addition, it allows models to interpret objects with basic geometric primitives. 

Our contributions are the following:

\begin{enumerate}
    \itemsep0em 
    \item We propose an end-to-end collaborative learning framework to leverage $3$D geometric information for compositional action recognition from egocentric RGB videos.
    \item We show that using superquadrics as the intermediate $3$D object representation is beneficial for $3$D hand pose estimation and interaction recognition. To the best of our knowledge, we are the first work to exploit superquadrics for both template-free object reconstruction and interaction recognition.
    \item We extend two egocentric hand-object datasets by introducing new compositional splits and investigate compositional action recognition where a subset of action verb and noun combinations do not exist during training.  
    \item We achieve state-of-the-art performance on two public datasets, H$2$O~\cite{kwon2021h2o} and FPHA~\cite{garcia2018first}, in both official and compositional settings.
\end{enumerate}

\section{Related Work}
Our work tackles the joint problem of $3$D hand-object reconstruction and action recognition from egocentric RGB videos. We first review the literature on reconstructing and recognising \emph{Hand-object interactions} from RGB inputs. Then, we provide a brief review on \emph{Compositional action recognition} and \emph{Superquadrics}.

\vspace{-0.1cm}
\paragraph{Reconstructing hand-object interactions.} 
While most existing research has primarily focused on single-hand~\cite{ge20193d,boukhayma20193d,baek2019pushing,simon2017hand,zimmermann2017learning} or objects~\cite{li2018deepim,zheng2022tp,wang2019densefusion,lepetit2004point,zheng2022tp,zheng2024georef} in isolation, recently there has been a surge in interest of joint understanding of hand-object pose estimation. As the problem of reconstructing both hand and object is extremely ill-posed due to heavy mutual occlusions, many works~\cite{cao2020reconstructing,tse2022s,liu2021semi,yang2021cpf,hampali2022keypoint,Yang_2022_CVPR} reduce this problem to $6$D pose estimation with instance-specific templates. Meanwhile, some previous efforts~\cite{hasson2019learning,tse2022collaborative,ye2022s,chen2022alignsdf,ye2023diffusion,chen2023gsdf} do not assume to have access to ground-truth object models at test time and follow a template-free paradigm. However, these approaches would fail on unseen object instances as they either lack geometrical prior or overfit to a limited number of training objects. Our work is a template-free approach where we utilise primitive shape information from estimated superquadrics, and therefore it is more generalisable to unseen object instances for understanding hand-object interaction.

\vspace{-0.1cm}
\paragraph{Recognising hand-object interactions.} 
Action recognition is one of the most actively researched areas in computer vision~\cite{jhuang2013towards,varol2017long,kantorov2014efficient} and significant progress has been made with the availability of large-scale datasets~\cite{sigurdsson2016hollywood,kay2017kinetics,karpathy2014large,grauman2022ego4d}. Here, we focus on methods~\cite{tekin2019h+,yang2020collaborative} that simultaneously estimate hand(-object) poses and interactions from egocentric videos. \citet{garcia2018first} presents the first egocentric dataset and shows that $3$D hand poses are beneficial for recognising actions. As the dataset contains visible magnetic sensors and does not include two-hand poses, a markerless dataset named H$2$O is developed to provide rich $3$D annotations for egocentric $3$D interaction recognition~\cite{kwon2021h2o}. This enables recent research to develop unified models
for understanding hand-object interactions. HTT~\cite{wen2023hierarchical} and H$2$OTR~\cite{cho2023transformer} are two closely related works that are based on a transformer architecture. HTT focuses on leveraging different temporal granularity information, while H$2$OTR exploits contact map for robust estimation. All of the above methods share a common approach by following the semantic relationship between hand poses and object labels to infer action. In contrast, we propose a collaborative learning framework which is designed explicitly to recognise interactions in a compositional fashion.

\vspace{-0.2cm}
\paragraph{Compositional action recognition.}
This task is designed to alleviate the problem of appearance bias by disjointing the combination of actions and objects between training and testing. STIN~\cite{materzynska2020something} models actions as transformation of geometric relations in both spatial and temporal domains.
This approach generalises well to most actions but fails when there are intrinsic state changes of objects. \citet{kim2020safcar} proposes to fuse RGB information with instance bounding boxes to capture more complex actions.
\citet{sun2021counterfactual} removes the appearance effect by counterfactual debiasing inference. 
While these methods have proven effective, they are primarily designed to leverage $2$D geometric information and do not fully explore the potential of $3$D geometric cues. As a result, their performance remains comparable to I$3$D~\cite{carreira2017quo}. In contrast, we focus on leveraging $3$D geometric cues for compositional action recognition.

\vspace{-0.2cm}
\paragraph{Superquadrics recovery.}
Superquadric is a well-studied computational primitive shape abstraction, offering a diverse range of shape representations including cuboids, ellipsoids, cylinders, octohedra, and other variations. It was first proposed to model complex objects in computer graphics~\cite{barr1981superquadrics}. \citet{solina1990recovery} presents a method for abstracting simple objects from range images using a single superquadric. Subsequently,~\citet{leonardis1997superquadrics,chevalier2003segmentation} extend to recover more complex objects with multiple superquadrics. Recently,~\citet{liu2022robust} proposes a probabilistic approach to improve robustness to outlier and fitting accuracy. To the best of our knowledge, we are the first to leverage superquadrics for action recognition and demonstrate its effectiveness in in-the-wild scenarios.
\section{Methodology}
\begin{figure*}[t]
\centering
\includegraphics[width=0.9\linewidth]{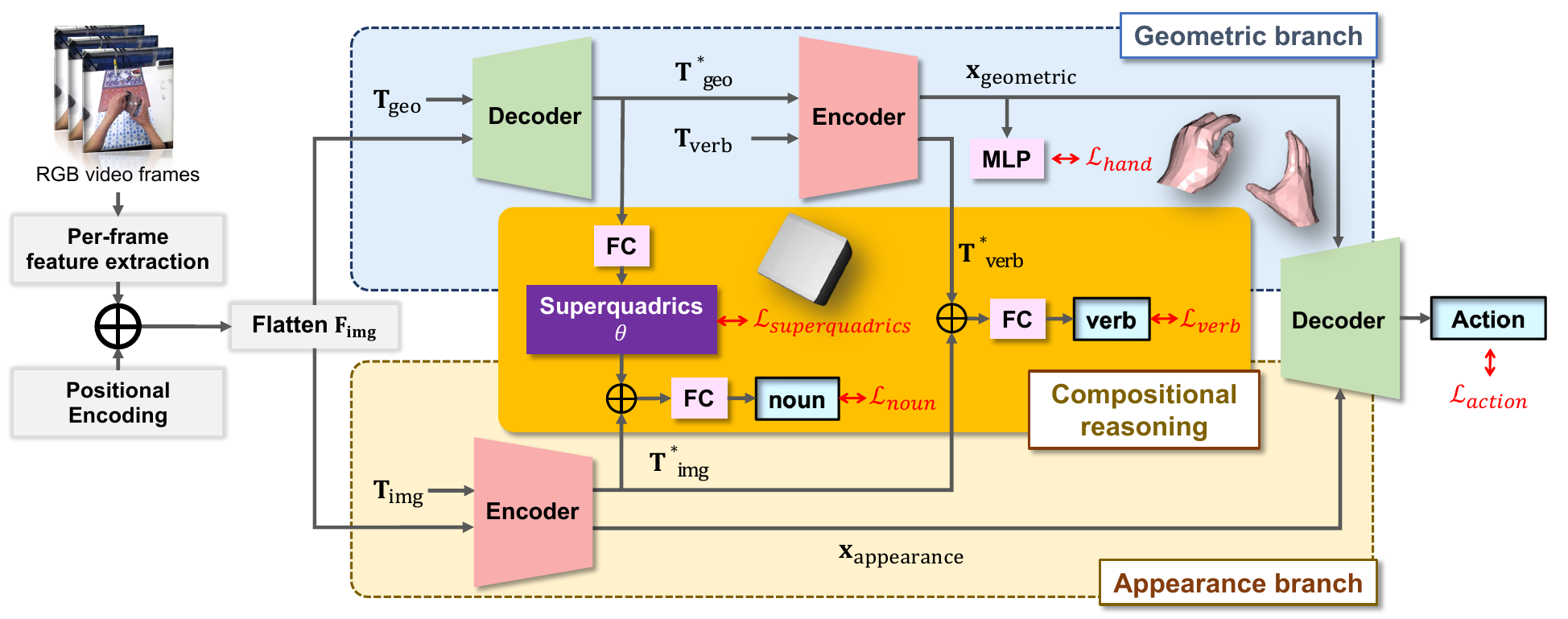}
\caption{
Overview of our approach. It first take RGB videos as input and produce per-frame spatial features $\bx$ using a CNN backbone. Then, the appearance branch (bottom) applies positional encoding to $\bx$ and combines it with a learnable token $\mathbf{T}_{\text{img}}$ before feeding into a Transformer encoder. Similarly, the geometric branch (top) extracts geometric features $\mathbf{T}_{\text{geo}}$ from flatten spatial features $\mathbf{F}_{\text{img}}$ using a Transformer decoder. The features from both branches are combined to predict superquadrics and object category. In addition, the geometric features are aggregated through another Transformer encoder to create global context-aware features between object shape and hand poses. The aggregated geometric features $\bx_{\text{geometric}}$ and verb token features $\mathbf{T}_{\text{verb}}$ from this encoder are used to predict hand pose and action verb. Finally, the action class is predicted by feeding $\bx_{\text{geometric}}$ into a cross-attention mechanism with the aggregated spatial representation $\bx_{\text{appearance}}$ through a Transformer decoder.
}
\label{fig:framework}
\end{figure*} 

Our training pipeline, as shown in \fig{framework}, takes a sequence of $T$ RGB frames $\mathbf{I}\in\mathbb{R}^{T\times256\times256\times3}$ of dynamic hands manipulating objects as input. We first obtain spatial features $\bx \in \mathbb{R}^{T\times d}$ by passing each frame into a ResNet-18 \cite{he2016deep} encoder where $d$ refers to feature dimensions. To enhance the interaction between visual and geometric cues, we propose a simple collaborative learning framework by leveraging the Transformer encoder and decoder~\cite{vaswani2017attention} as basic building blocks. Specifically, we design a two-branch network where the appearance branch extracts video features and the geometric branch aims to recover $3$D hand-object geometric information. With such design, we can model appearance and geometric information at different temporal granularity. In addition, we explicitly model the compositionality of the interaction by decomposing the action class into a verb-and-noun pair. Finally, we combine video appearance and geometric representations for recognising egocentric hand-object interactions.

\vspace{-0.1cm}
\subsection{Appearance Branch} \label{sec:appearance}

Given image features $\bx$, we concatenate with a learnable token $\mathbf{T}_{\text{img}} \in \mathbb{R}^{d}$ and apply positional encoding before feeding them into a Transformer encoder. The encoder models the relationships of different spatial regions and $\mathbf{T}_{\text{img}}$ through self-attentions. The learnt token $\mathbf{T}^{*}_{\text{img}}$ captures essential global contexts from backbone representation which are used for compositional reasoning. In addition, the encoder outputs aggregated spatial representation $\bx_{\text{appearance}} \in \mathbb{R}^{T\times d}$ which is later used for interaction recognition.


\vspace{-0.1cm}
\subsection{Geometric Branch} \label{sec:geometric}
As estimating the poses of hand and object requires more local or nearby frames, we divide the video sequence into $N$ consecutive segments. Specifically, we follow~\citet{wen2023hierarchical} and use a shifting window strategy with window size $t$, \ie $N=T/t$. The frames beyond sequence length $T$ are padded but masked out from attention computation. Readers are referred to~\citet{wen2023hierarchical} for more details.

Instead of aiming to reconstruct the hand and the manipulated object simultaneously, we first estimate the shape and poses of the manipulated object and leverage this geometric information to predict hand poses. The reason behind this approach is that joint estimation poses a significantly harder problem: First, self-occlusion and self-similarity between the joints of two hands are unique problems in interacting hands. Second, when interacting with objects, hands and objects often exhibit even greater occlusions. This problem is further amplified under egocentric view setting due to large degree of erratic camera motions. 

In addition, existing work which relies on $2$D/$3$D bounding boxes has limitations in accurately representing the shape and movement of objects. However, at the same time, it remains challenging to accurately reconstruct unseen objects from RGB images. Therefore, we propose to use superquadrics as a new object representation for improving action recognition. In the following, we present the \emph{preliminaries} of superquadrics and detail our two-stage approach consisting \emph{superquadrics decoder} and \emph{hand pose estimator}.


\begin{figure}
\centering
\includegraphics[width=0.9\linewidth]{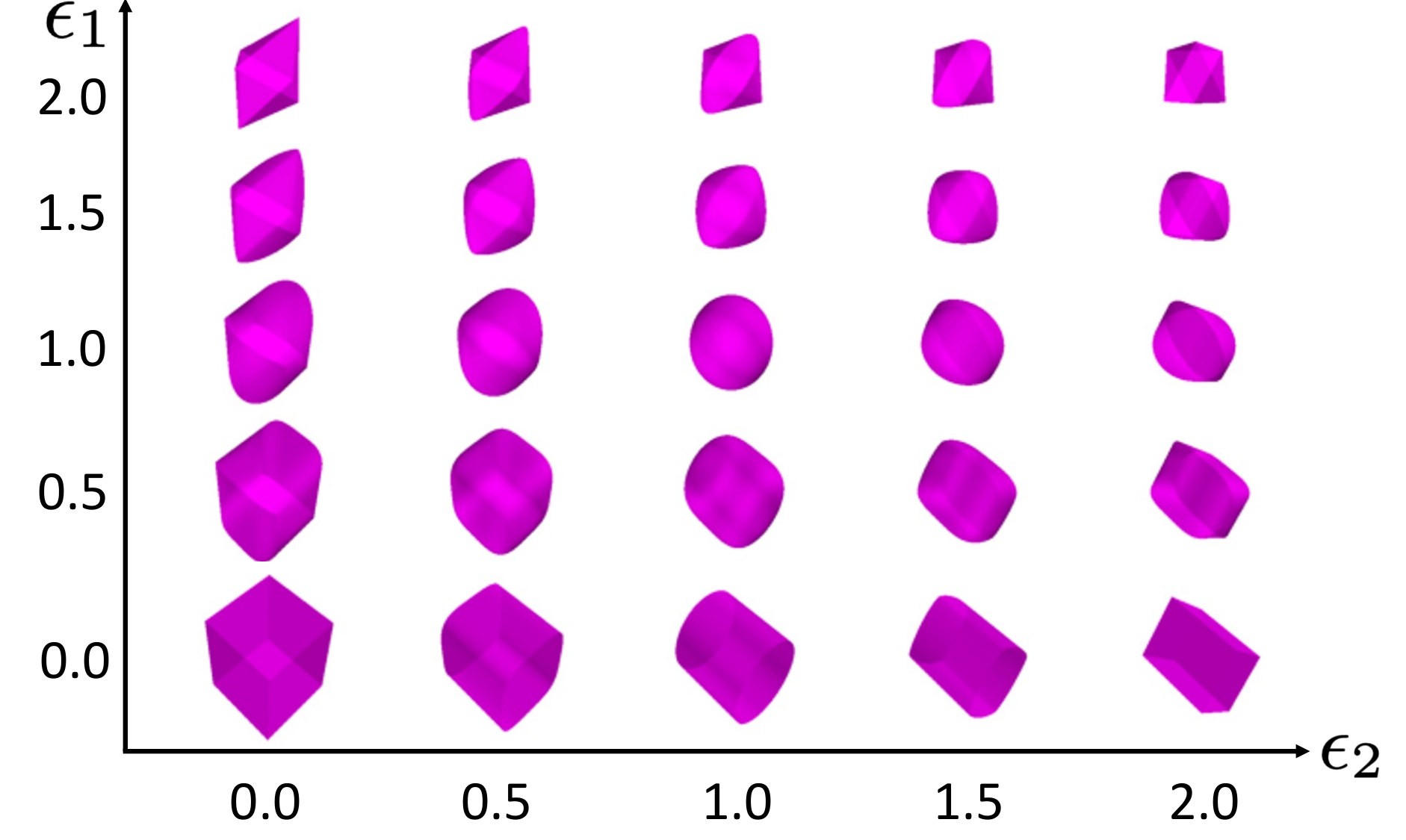}
\caption{Qualitative examples of convex superquadrics. We show that superquadrics can model diverse objects by varying the shape parameters, $\epsilon_1$ (y-axis) and $\epsilon_2$ (x-axis).
}
\label{fig:sq}
\end{figure}

\paragraph{Preliminaries.} As shown in \fig{sq}, superquadrics are a family of geometric primitives, \ie cuboids, cylinders, ellipsoids, octahedra and their intermediates, which can be defined by an implicit function $f(\cdot)$~\cite{barr1981superquadrics}:
\begin{small}
\begin{align}\label{eq:sq}
    f(\mathbf{p}) = \Biggl( \biggl( \frac{x}{a_x} \biggr)^{\frac{2}{\epsilon_2}} + \biggl( \frac{y}{a_y} \biggr)^{\frac{2}{\epsilon_2}} \Biggr)^{\frac{\epsilon_2}{\epsilon_1}} +  \biggl( \frac{z}{a_z} \biggr)^{\frac{2}{\epsilon_1}} = 1,
\end{align}
\end{small}
where points $\mathbf{p} = [x,y,z] \in \mathbb{R}^{3}$ satisfying \eq{sq} form the surface of a superquadric. It can be encoded using $5$ parameters: shape parameters $\epsilon_1,\epsilon_2 \in [0,2] \subset \mathbb{R}$ and scale parameters $a_x,a_y,a_z \in \mathbb{R}_{>0}$. While the shape parameters can exceed $2$ and result in non-convex shapes, we limit them within the convex region in this paper. We can now fully parameterise a superquadric by including the Euclidean transformation $g \in SE(3)$, \ie $g = [\mathbf{R} \in SO(3), \mathbf{t} \in \mathbb{R}^3]$.

\begin{figure}
\centering
\includegraphics[width=0.9\linewidth]{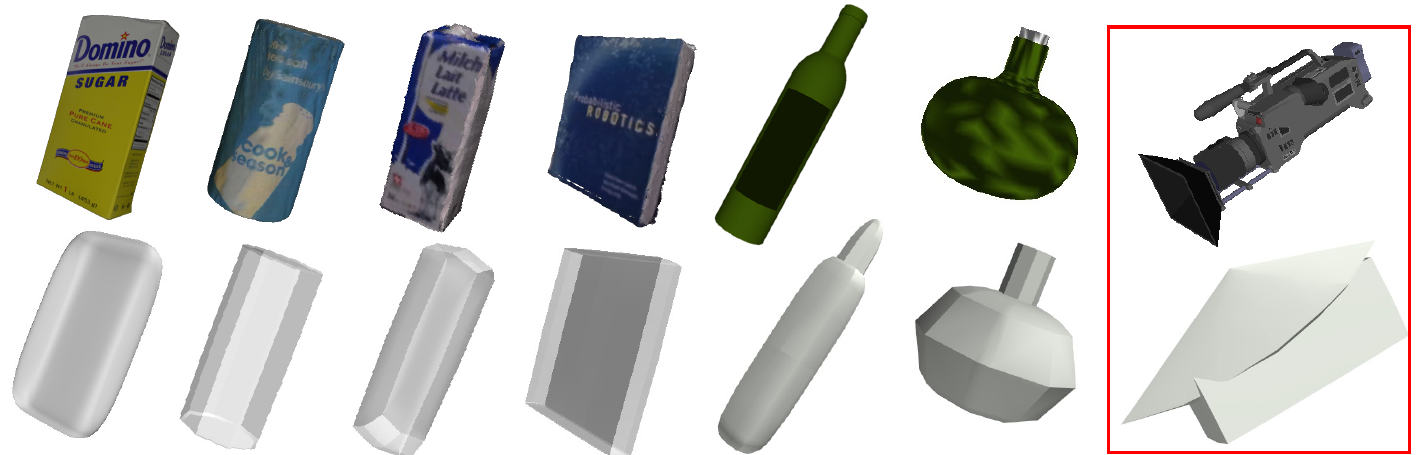}
\caption{Qualitative examples of superquadrics. We extract superquadrics from everyday objects obtained from \emph{YCB}~\cite{calli2015ycb}, \emph{ShapeNet}~\cite{shapenet2015}, \emph{FPHA}~\cite{garcia2018first} and \emph{H$2$O}~\cite{kwon2021h2o} datasets. We show that superquadrics have sufficient expressiveness to represent everyday objects. We also present an example failure case in the red box.
}
\label{fig:shapenet}
\end{figure}
\paragraph{Superquadrics decoder.} We use superquadrics $\boldsymbol{\theta}$ to be the intermediate $3$D object representation, as it offers a compact way to represent a wide range of geometric primitives, \ie $\boldsymbol{\theta} = \{\epsilon_1,\epsilon_2,a_x,a_y,a_z,g \} \in \mathbb{R}^{11}$. As shown in \fig{shapenet}, superquadrics can provide sufficient expressiveness to reasonably model a diverse range of everyday objects. To this end, we extract geometric features $\mathbf{T}_{\text{geo}} \in \mathbb{R}^{N\times d}$ from the segmented spatial features using a Transformer decoder. We train a fully-connected layer to predict $\boldsymbol{\theta}$ from the learnt flatten $\mathbf{T}^{*}_{\text{geo}}$ by minimising the L$1$ loss $\loss_{\text{superquadrics}}$. The main motivation to use Transformer decoder here instead of encoder is to leverage the autoregressive self-attention mechanism for modelling objects with multiple superquadrics. 
In addition, having separate encoder and decoder for this branch enables pre-training on large-scale object datasets that lack hand pose annotations.

\paragraph{Hand pose estimator.} Given the learnt geometric features $\mathbf{T}^{*}_{\text{geo}}$, we concatenate with a learnable verb token $\mathbf{T}_{\text{verb}}$ and use self-attention in a Transformer encoder to create global context-aware features between object shape and hand poses. This encoder outputs aggregated geometric features $\bx_{\text{geometric}} \in \mathbb{R}^{N\times d}$ and verb token features $\mathbf{T}^{*}_{\text{verb}} \in \mathbb{R}^{d}$.

We map the aggregated geometric features $\bx_{\text{geometric}}$ to hand pose space by a $3$-layer MLP and use MANO joint angles~\cite{romero2017embodied} for hand pose representation. Specifically, we estimate $16$ $3$D joint angles under the hand kinematic tree and MANO hand shape parameter per hand. Then, we can compute the $21$ root-relative $3$D joint locations of each hand by using the predicted joint angles and hand shape parameters. They are learned by minimising L$1$ loss $\loss_{\text{hand}}$.

\vspace{-0.1cm}
\subsection{Compositional Reasoning} \label{sec:compositional}
In the following, we describe how we leverage the compositional nature of actions by exploiting the action class as verb-noun pair with $3$D geometric cues, \ie superquadrics $\boldsymbol{\theta}$ and geometric-aware verb token $\mathbf{T}^{*}_{\text{verb}}$.

\paragraph{Object category predictor.} We first predict the category of the manipulated object as it corresponds to the noun of an action. To achieve that, we leverage the basic primitive geometric information from superquadrics $\boldsymbol{\theta}$ as object shape provides strong signals to estimating object category. More specifically, we predict the classification probability vector for object category by linearly projecting the concatenation of superquadrics $\boldsymbol{\theta}$ and $\mathbf{T}^{*}_{\text{img}}$. We supervise this linear layer by minimising the cross-entropy loss $\loss_{\text{noun}}$.

\paragraph{Verb predictor.} Similarly, we predict action verb by feeding the concatenation of $\mathbf{T}^{*}_{\text{verb}}$ and $\mathbf{T}^{*}_{\text{img}}$ to a linear layer. It is also trained by minimising the cross-entropy loss $\loss_{\text{verb}}$.

\paragraph{Discussion.} The key idea for concatenating with $\mathbf{T}^{*}_{\text{img}}$ is to allow action verb and noun to interact with the appearance branch. Also, it generates loss gradients for both branches to develop a collaborative learning relationship. In addition, the motivation for estimating superquadrics first in the geometric branch is based on the fact that the human visual system flavours abstracting scenes into canonical parts for better perceptual understanding~\cite{liu2022robust}. This enables robust action recognition using basic geometric primitives instead of relying on accurate point-wise estimation. In summary, our design targets the problem of recognising a seen action when facing new objects by enabling the network to capture the compositionality of an action explicitly.

\vspace{-0.1cm}
\subsection{Interaction Recognition} \label{sec:interaction}

Besides explicitly modelling the compositionality of actions, our proposed framework can easily combine with any video-level appearance representation. The impact of appearance features can be two-fold: 1) The presence of appearance features can be particularly beneficial for action classes that lack prominent inter-object dynamics~\cite{materzynska2020something}. 2) Conversely, appearance bias can inhibit the model learning ability by making strong correlations on spatial appearance rather than temporal or geometric transformations~\cite{sun2021counterfactual}. To overcome the limitations of existing methods that can only accept or reject appearance information as a whole, we propose to use a Transformer decoder for recognising interaction. There are two key motivations for this design: 1) Transformer decoder has skip connections by design so it can attend segments of appearance features. 2) It can take different temporal granularity from both branches as inputs to make action predictions. Specifically, this decoder takes the aggregated geometric and spatial features, \ie $\bx_{\text{geometric}}$ and $\bx_{\text{appearance}}$ as input and extracts relevant image features through cross-attention between geometric features. The vector output of this decoder is fed to a $3$-layer MLP classifier of width and is supervised with cross-entropy loss $\loss_{\text{action}}$. We investigate and analyse our design choices and provide details for the architecture in the supplementary materials.

\section{Experiments} \label{sec:experiment}
\vspace{-0.1cm}
\paragraph{Implementation details.}
We train our model with the Adam optimiser~\cite{kingma2014adam} using an initial learning rate of $3\times10^{-5}$ which halves in every $15$ epochs. 
We keep the relative weights between different losses and normalise them for all experiments.
We use ResNet~\cite{he2016deep} pre-trained on ImageNet~\cite{russakovsky2015imagenet} for our backbone. For all Transformer encoders and decoders, we use $2$ encoding/decoding layers where each layer has $8$ attention heads. We use the fixed sine/cosine functions for positional encoding and add layer normalisation before the attention and feed-forward computations~\cite{vaswani2017attention}. We follow~\citet{wen2023hierarchical} by setting $T=128$, $t=16$, $d=512$ and training for $45$ epochs with batch size of $2$. Our final loss $\loss_{\text{final}}$ is defined as:
\vspace{-0.1cm}
\begin{equation}
    \loss_{\text{final}} = \loss_{\text{superquadrics}}+\loss_{\text{hand}}
    + \loss_{\text{noun}}+\loss_{\text{verb}}+\loss_{\text{action}}.
\end{equation}
\vspace{-0.1cm}
\paragraph{Datasets.} We conduct experiments on $3$ interacting hand-object datasets and detail below.

\emph{ObMan}~\cite{hasson2019learning} is a synthetic dataset which was produced by rendering hand meshes with selected objects from ShapeNet~\cite{chang2015shapenet} dataset. It captures 8 object categories of everyday objects (\eg bottles, cans and jars) and results in a total of 2,772 meshes. We precomputed superquadrics for all object meshes using the EMS algorithm~\cite{liu2022robust} and pretrained the geometric branch on \emph{ObMan} before training on other real datasets. We observed consistent improvements over training directly on real data as the number of objects in hand-object interaction dataset is very limited. 

\emph{First-person hand benchmark (FPHA)}~\cite{garcia2018first} is a real dataset which records egocentric videos on diverse hand-object interactions. It captures $6$ subjects performing $45$ actions by interacting with $26$ objects, \ie juice bottle, liquid soap, milk and salt. 
We evaluate on the \textit{action split} where all subjects and actions are present in both training and testing. This split consists of $600$ and $575$ videos for training and testing, respectively.

\emph{H2O}~\cite{kwon2021h2o} is a recent real dataset which provides markerless $3$D annotations for two hands and the $6$D pose of manipulated objects. It is the first unified dataset for egocentric interaction recognition with rich $3$D annotations of $4$ subjects performing $36$ actions on $8$ objects. We follow~\citet{wen2023hierarchical,kwon2021h2o} and use the sequences of egocentric view for training and testing. Specifically, the training split consists of $569$ videos from the first $3$ subjects, while the testing split includes $242$ videos from the remaining unseen subjects.

\newcolumntype{C}{>{\centering\arraybackslash}X}
\begin{table*}
\begin{center}
\scalebox{0.9}{
\begin{tabular}{l | ccc | ccc }
\toprule
 & \multicolumn{3}{c|}{\emph{H$2$O}} & \multicolumn{3}{c}{\emph{FPHA}} \\
Method & \small{Top-$1$ accuracy ($\%$) $\uparrow$} & \small{Hand (L/R) $\downarrow$} & \small{Obj. $\downarrow$} & \small{Top-$1$ accuracy ($\%$) $\uparrow$} & \small{Hand $\downarrow$}  & \small{Obj. $\downarrow$} \\
\midrule
C$2$D~\citep{wang2018non} & 70.66 & - & - & - & - & - \\
I$3$D~\citep{carreira2017quo} & 75.21 & - & - & - & - & - \\
SlowFast~\citep{feichtenhofer2019slowfast} & 77.69 & - & - & - & - & - \\
MViTv$2$~\citep{li2022mvitv2} & 90.08 & - & - & 98.45 & - & - \\
\midrule
H+O~\citep{tekin2019h+} & 68.88 & 41.4/38.9 & 50.4 & 82.43 & 15.8 & 24.9  \\
H$2$O~\citep{kwon2021h2o}  & 79.25 & 41.5/37.2 & 47.9 & -  & -& -\\
HTT~\citep{wen2023hierarchical} & 86.36 & 35.0/36.1 & - & 94.09 & -& -\\
H$2$OTR~\citep{cho2023transformer} & 90.90 & \textbf{24.4/25.8} & 45.2 & 98.4 & 15.0 & 21.0\\
\midrule
Ours & \textbf{92.25} & 28.9/30.2 & \textbf{43.5} & \textbf{98.74} & \textbf{13.6} & \textbf{20.1}\\
\bottomrule
\end{tabular}%
}   
\end{center}
\caption{Error rates of action recognition and pose estimation on \emph{H2O} and \emph{FPHA}. We report the top-1 accuracy for classification and MEPE in $mm$ for hand (left/right) and object error. Our proposed method performs competitively without known object templates at inference.
}
\label{tbl:ch5:action}
\end{table*} 
\begin{figure}[ht]
\centering
\includegraphics[width=1\linewidth]{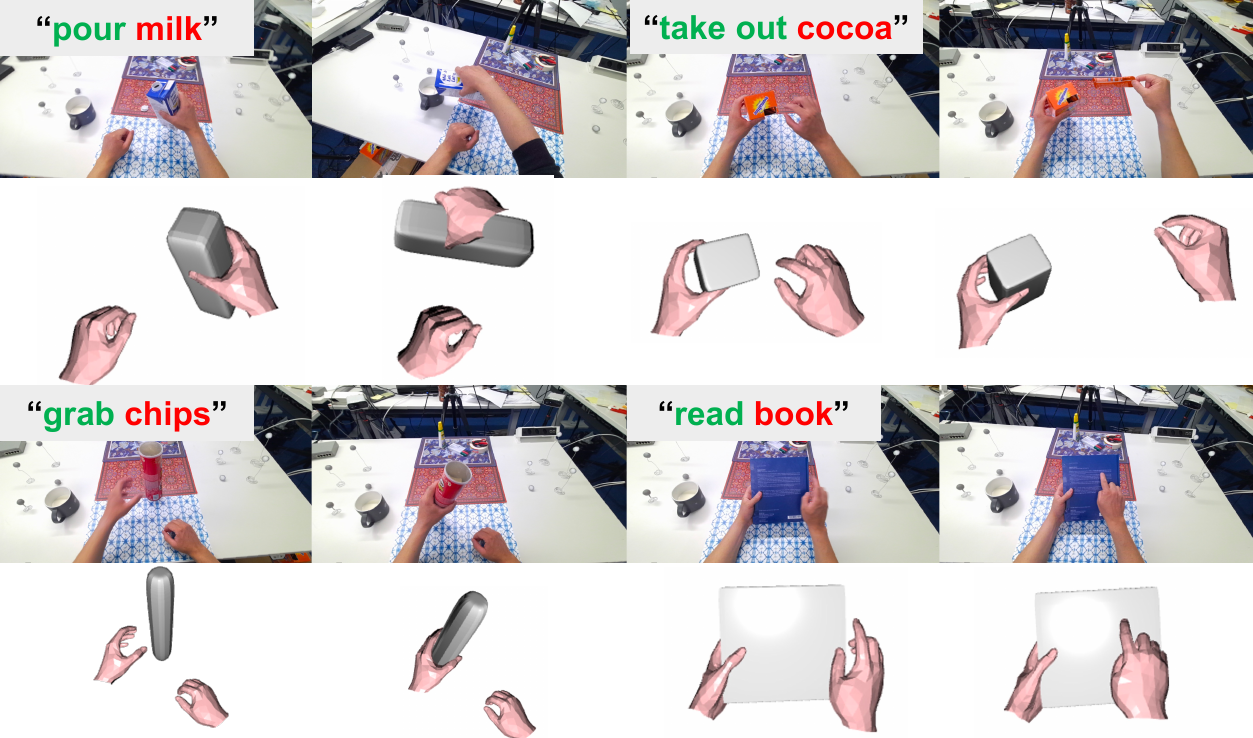}
\caption{Qualitative examples on \emph{H$2$O}. We show that our model can recover plausible interaction across different object categories and hand-object configurations without object templates. We present additional qualitative examples and failure cases in the supplementary.}
\label{fig:qualitative}
\end{figure} 
\paragraph{Baselines.}
We compare our method against MViTv$2$~\cite{li2022mvitv2}, HTT~\cite{wen2023hierarchical} and H2OTR~\cite{cho2023transformer}. MViTv$2$ is widely adopted for video recognition tasks and can serve as a strong appearance-based baseline. We train the base variant of MViTv$2$ with weights pretrained on Kinetics-$400$ dataset~\cite{kay2017kinetics} using the PySlowFast library~\cite{fan2020pyslowfast}. For the pose-based baselines, we consider HTT and H2OTR as they are recent methods that achieves state-of-the-art performance on both FPHA and H$2$O datasets. We also consider two Transformer-based baselines which do not contain compositional reasoning and superquadrics, respectively. We provide more details about the baselines in the supplementary materials. They are useful for understanding the importance of superquadrics and compositional reasoning for recognising interactions with unseen objects.

\paragraph{Evaluation metrics.} We report the Mean End-Point Error (MEPE) in $mm$ to evaluate \emph{pose estimation}. MEPE measures the mean Euclidean distances between predictions and ground-truths. We also report the top-$1$ classification accuracy for \emph{action recognition}.

\newcolumntype{C}{>{\centering\arraybackslash}X}
\begin{table*}[h]
\begin{center}
\resizebox{0.99\linewidth}{!}{
\begin{tabular}{l | c c | c c | c c | c c | c c}
\toprule
 & \multicolumn{6}{c|}{\emph{H$2$O}} & \multicolumn{4}{c}{\emph{FPHA}} \\
 \midrule
 Method & $\mathcal{S}_0 (\%) \uparrow$ & Hand $\downarrow$ & $\mathcal{S}_1 (\%) \uparrow$ & Hand $\downarrow$ & $\mathcal{S}_2 (\%) \uparrow$ & Hand $\downarrow$ & $\mathcal{S}_0 (\%) \uparrow$ & Hand $\downarrow$ & $\mathcal{S}_1 (\%) \uparrow$ & Hand $\downarrow$ \\
\midrule
MViTv$2$~\cite{li2022mvitv2} &
\small{90.08} & - &
\small{60.64$\pm$2.3} & - &
\small{52.32$\pm$5.7} & - &
\small{98.45} & - &
\small{69.02$\pm$1.8} & - \\

HTT~\cite{wen2023hierarchical} & 
\small{86.36} & \small{35.6} &
\small{71.13$\pm$2.7} & \small{37.4$\pm$2.3} &
\small{59.88$\pm$2.6} & \small{41.5$\pm$1.7} &
\small{94.09} & \small{15.8} &
\small{74.21$\pm$1.8} & \small{18.8$\pm$1.2}\\ 

H2OTR~\cite{cho2023transformer} &
\small{90.90} & \small{25.1} &
\small{73.54$\pm$67} & \small{35.6$\pm$3.7} &
\small{61.78$\pm$6.1} & \small{40.2$\pm$5.2} &
\small{98.4} & \small{15} &
\small{76.52$\pm$9.1} & \small{18.3$\pm$4.2}\\ 

\midrule
Ours  & 
\small{\textbf{92.25}} & \small{29.6} &
\small{\textbf{80.59$\pm$1.6}} & \small{\textbf{31.8$\pm$2.1}}  &
\small{\textbf{69.93$\pm$2.5}} & \small{\textbf{33.4$\pm$1.5}} &
\small{\textbf{98.74}} & \small{\textbf{13.6}}  &
\small{\textbf{85.80$\pm$1.4}} & \small{\textbf{13.9$\pm$0.9}} \\
\bottomrule
\end{tabular}
}
\end{center}
\caption{
Error rates of compositional action recognition \emph{H$2$O} and \emph{FPHA}. We report classification accuracy in $\%$ and hand error in $mm$ for the official split $\mathcal{S}_0$ and two additional compositional split settings, $\mathcal{S}_1$ and $\mathcal{S}_2$. The results are reported as the mean and standard deviation of the performance metric of interest, providing a comprehensive understanding of the model's performance across different split settings.
}
\label{tbl:composition}
\end{table*}
\newcolumntype{C}{>{\centering\arraybackslash}X}
\begin{table*}[h]
\begin{center}
\resizebox{0.9\linewidth}{!}{
\begin{tabular}{l | c c c | c | c }
\toprule
 Method & \small{Verb$(\%) \uparrow$} & \small{Noun$(\%) \uparrow$} & \small{Verb+Noun$(\%) \uparrow$} & \small{Top-$1$$(\%) \uparrow$} & \small{Hand$(mm) \downarrow$}\\
\midrule
w/o geometric branch & - & - & - & 78.91 & - \\ 
w/o compositional reasoning & - & - & - & 81.82 & 38.91 \\
w/o verb classifier & - & 85.61 & - & 83.45 &  37.15 \\
w/o noun classifier & 86.24 & - & - & 83.96 & 37.62 \\
w/o interaction decoder & 88.07 & 90.04 & 87.2 & - & 32.67 \\
w/o superquadrics & 90.18 & 91.56 & 90.01 & 89.85 & 31.85\\
\midrule
Ours  & \textbf{92.23} & \textbf{96.89} & \textbf{92.25} & \textbf{90.0} & \textbf{29.6} \\
\bottomrule
\end{tabular}}
\end{center}
\caption{Ablation study of model architecture design on \emph{H$2$O}. We report top-$1$ classification accuracy $(\%)$ and MEPE for hand error in $mm$. 
In addition,
we include classification predictions for verb and noun from compositional reasoning.
}
\label{tbl:ablations}
\end{table*}

\vspace{-0.1cm}
\subsection{Quantitative Comparison}
\vspace{-0.1cm}

\paragraph{Comparison with state-of-the-art.} We report quantitative comparisons with the state-of-the-art methods on \emph{H$2$O} and \emph{FPHA} datasets in Table~\ref{tbl:ch5:action}. We split the table into two sections: appearance-based~\cite{wang2018non,carreira2017quo,feichtenhofer2019slowfast,li2022mvitv2} and geometric~\cite{tekin2019h+,yang2020collaborative,kwon2021h2o,wen2023hierarchical,cho2023transformer} methods for clear comparison. By considering the performances of action recognition, the appearance-based baseline MViTv$2$ performs competitively with the state-of-the-art geometric method H$2$OTR~\cite{cho2023transformer}. It raises an immediate question as to when $3$D geometric cues be beneficial to recognising interaction as collecting ground-truth contact maps or other $3$D annotations are non-trivial~\cite{brahmbhatt2019contactdb,tse2022s,kwon2021h2o}. We will address this question in the following paragraph. Nonetheless, we demonstrate the effectiveness of explicit compositional reasoning with superquadrics by outperforming all methods. On the other hand, for pose estimations, H$2$OTR is a Transformer-based framework which achieves state-of-the-art accuracy for hand pose estimation. 
However, all of the compared methods cannot recover dense object geometries without manual selection of object models at test time.
In contrast, our method performs competitively without known object templates and outperforms all methods on object pose estimation. We attribute this to the fact that superquadrics can provide dense $3$D geometric information about the manipulated object, whereas $2$D or $3$D bounding boxes have limitations on representing object shape and movement. We show qualitative results on \emph{H$2$O} dataset in \fig{qualitative}.

\paragraph{Compositional action recognition.} We further evaluate on the compositional recognition task in Table~\ref{tbl:composition}. Following~\citet{materzynska2020something}, we first create new splits for the task of compositional action recognition by extending existing egocentric hand-object datasets. Specifically, we remove the sequences that contain the predefined object category from the train split, such that the combinations of a verb (action) and nouns do not overlap in the testing set. To gain a deeper understanding of the model generalisation ability on unseen objects, we evaluate the model using $N_{\text{obj}}-$fold cross validation where $N_{\text{obj}}$ refers to the number of total objects presented in the dataset. We keep the original testing splits (named $\mathcal{S}_0$) to illustrate the difficulty of this compositional task. We further experiment on a more challenging split where two object categories are randomly removed in the \emph{H$2$O} dataset. We name the base splits by $\mathcal{S}_1$ and the more difficult splits where additional verb-nouns combinations are removed from training by $\mathcal{S}_2$. We report the mean and the standard deviation of top-$1$ classification accuracy for all experiments. We present the main results in Table~\ref{tbl:composition}, while full results are presented in the supplementary.

As shown in Table~\ref{tbl:composition}, our proposed collaborative learning framework consistently outperforms both the appearance and geometric baselines, \ie MViTv$2$, HTT and H2OTR. We find that the performance of MViTv$2$ drastically drops by $29.44\%$ and $37.76\%$, in $\mathcal{S}_1$ and $\mathcal{S}_2$ of H$2$O respectively. These results are in line with previous studies~\cite{materzynska2020something,zhou2018temporal,sun2021counterfactual} where deep architectures tend to overfit the object appearance. By adding geometric cues in HTT and H2OTR, we observe a small performance gain by an average of $7.75\%$ and $9.95\%$ on both compositional splits, respectively. We further evaluate hand pose estimation accuracy under this compositional setting. Similarly, we report the mean and the standard deviation of MEPE in $mm$ for all experiments. By comparing with HTT, we achieve state-of-the-art performance in hand pose estimation with the advantage of object reconstruction using superquadrics. Our strong performances across all settings demonstrate the importance of explicit reasoning about interactions with $3$D geometric information.

\vspace{-0.1cm}
\subsection{Ablations}
To motivate our design choices, we perform additional qualitative evaluation of our method with various key components disabled in Table~\ref{tbl:ablations}. We evaluate the effectiveness of our collaborative learning framework by experimenting on a single appearance branch (row $1$) and a two-branch network without gradient flow (row $2$), which yields the lowest performance in \emph{H$2$O}. Then, we observe significant performance drops by removing either one of the inter-branch classifiers (row $3,4$). These results demonstrate the effectiveness of our collaborative learning framework which encourages information sharing between two-branches. Further, we are interested in finding out whether incorporating the interaction decoder is important as action class can be obtained by combining verb and noun predictions. We show that the interaction decoder can bring performance gain for verb and noun classifiers by additional supervision in row $5$. Finally, we show that superquadric predictions (\ie by removing the purple block in \fig{framework}) can push the limit to the new state-of-the-art in all metrics.


\section{Conclusion}
We showed that, by explicitly leveraging $3$D geometric information, we could recognise actions performed on unseen objects much more accurately than existing state-of-the-arts. We also demonstrated that superquadrics as a new object representation for action recognition to be effective. We validated our approach by extending existing datasets with compositional splits and achieved state-of-the-art performance. We plan to investigate more complex interactions with articulated objects in the future.

\paragraph{Limitations.} Our approach relies on the expressiveness of superquadrics. First, we found that multi-superquadrics recovery is necessary to model more complex shapes, where preliminary point cloud segmentation can be helpful. Second, it remains challenging to capture non-convex everyday objects such as cups, papers and clothes. Third, we found superquadrics recovery relies on the quality of the object template. We show in supplementary that a broken object model can heavily degrade the accuracy of superquadrics recovery. We plan to extend the vocabulary of superquadrics by incorporating deformation fields and articulation joints.

\section{Acknowledgments}
This research was supported by the MSIT (Ministry of Science and ICT), Korea, under the ITRC (Information Technology Research Center) support programme (IITP-2024-2020-0-01789) supervised by the IITP (Institute for Information \& Communications Technology Planning \& Evaluation). This work is supported in part by the National Natural Science Foundation of China under Grant 62203184.
\section{Appendix}
In this supplemental document, we provide:
\begin{itemize}
    \setlength{\itemsep}{0pt}%
    \setlength{\parskip}{0pt}%
    \item implementation details for networks;
    \item additional results and analysis;
    \item additional qualitative examples;
    \item broader impacts.
\end{itemize}

\subsection{Implementation Details} \label{sec:implementation}

\paragraph{Transformer encoder.}
The encoder consists of $2$ identical layers. Each of the layers is composed of a multi-head self attention mechanism and a positional-wise fully connected feed-forward network. Following~\citet{vaswani2017attention}, we employ residual connections around both sub-layers, followed by layer normalisation. The encoder is used to aggregate input features by performing self-attention.

\paragraph{Transformer decoder.}
Similarly, the decoder consists of $2$ identical layers. In addition to the two sub-layers in encoder layer, the decoder layer has another multi-head attention mechanism for the secondary input. It also consists of residual connections around each of the sub-layers, followed by layer normalisation. The decoder is used to extract features by performing self-attentions on primary inputs and cross-attentions together with secondary inputs. We refer readers to~\cite{vaswani2017attention} for additional implementation details.

\paragraph{MViT baseline.} We use the PySlowFast library~\cite{fan2020pyslowfast} for performing action recognition comparisons with the state-of-the-art appearance-based methods, \ie C$2$D~\cite{wang2018non}, I3D~\cite{carreira2017quo}, SlowFast~\cite{feichtenhofer2019slowfast} and MViT~\cite{fan2021multiscale}. As we found that MViT performs significantly better than the rest of appearance-based methods in our preliminary experiments, we chose MViT to serve as a strong image-based learning baseline. For all experiments, we use MViTv2-B model~\cite{fan2021multiscale} pre-trained on Kinetics 400~\cite{kay2017kinetics}. We found that utilising a pre-trained model is essential as we observed a significant performance drop to less than $25\%$ when training from scratch without any pre-training. MViT-B consists of $4$ scale stages where each of them have several transformer blocks of consistent channel dimension. The model initially projects the dimension of the input channel to $96$ with patch kernel shape of $(3, 7, 7)$. We set patch stride and padding to be $(2, 4, 4)$ and $(1, 3, 3)$, respectively. The resulting sequence is reduced by a factor of $4$ for each additional stage. The number of head for multi-head pooling attention (MHPA) is set to $1$ at initial stage and increases with channel dimension. The output dimension of MLP at each transition is increased by $2$ and MHPA pools $\mathcal{Q}$ tensors with stride $(1, 2, 2)$ as the input for the next stage. The K, V pooling in MPHA employs stride at $(1, 8, 8)$ and adaptively adjusts with respect to the scale across stages to ensure consistent scaling of the K, V tensors in all blocks. We refer readers to~\citet{fan2021multiscale} for more details.

\subsection{Additional Analysis} \label{sec:additional_analysis}

\subsubsection{Compositional action recognition}
\paragraph{FPHA.}
We present full results for compositional action recognition on \emph{FPHA} dataset in Table~\ref{tbl:fpha_full}.

\newcolumntype{C}{>{\centering\arraybackslash}X}
\begin{table*}[h]
\begin{center}
\resizebox{1\linewidth}{!}{
\begin{tabularx}{\linewidth}{l | C C | C C | C C}
\toprule
  & \multicolumn{2}{c|}{MViTv$2$~\cite{li2022mvitv2}} & \multicolumn{2}{c|}{HTT~\cite{wen2023hierarchical}} & \multicolumn{2}{c}{Ours} \\
 \midrule
 Unseen obj. & Top-$1$ ($\%$) $\uparrow$ & Hand $\downarrow$ & Top-$1$ ($\%$) $\uparrow$ & Hand $\downarrow$ & Top-$1$ ($\%$) $\uparrow$ & Hand $\downarrow$ \\
\midrule
Juice bottle &
71.10 & - &
76.44 & 19.4 &
85.41 & 14.6  \\

Milk & 
69.81 & - &
72.38 & 18.3 &
87.45 & 13.9  \\

Salt & 
67.92 & - &
74.85 & 20.1 &
86.23 & 12.6  \\

Liquid soap & 
67.23 & - &
73.15 & 17.5 &
84.12 & 13.9  \\

\bottomrule
\end{tabularx}}
\end{center}
\caption{
Error rates of compositional action recognition on \emph{FPHA}. We report classification accuracy in $\%$ and hand error in $mm$. 
}
\label{tbl:fpha_full}
\end{table*}

\paragraph{H2O.}
We present full results for compositional action recognition on \emph{H2O} dataset for splits $\mathcal{S}_1$ and $\mathcal{S}_2$ in Table~\ref{tbl:H2O_full_s1} and Table~\ref{tbl:H2O_full_s2}, respectively.

\newcolumntype{C}{>{\centering\arraybackslash}X}
\begin{table*}[h]
\begin{center}
\resizebox{1\linewidth}{!}{
\begin{tabular}{l | c | c c c | c c c}
\toprule
  & MViTv$2$~\cite{li2022mvitv2} & \multicolumn{3}{c|}{HTT~\cite{wen2023hierarchical}} & \multicolumn{3}{c}{Ours} \\
 \midrule
 Unseen obj. & Top-$1$ ($\%$) $\uparrow$ & Top-$1$ ($\%$) $\uparrow$ & Hand (L) $\downarrow$ & Hand (R) $\downarrow$ & Top-$1$ ($\%$) $\uparrow$ & Hand (L) $\downarrow$ & Hand (R) $\downarrow$  \\
\midrule
Book &
64.05 & 
69.55 & 41.06 & 41.55 &
79.72 & 35.95 & 36.21 \\

Espresso &
62.81 & 
77.41 & 37.19 &  37.09 &
81.65 & 31.21 & 33.59 \\

Lotion &
56.61 & 
72.38 & 36.52 & 36.92 &
81.81 & 30.68 & 33.56 \\

Spray &
60.74 & 
69.46 & 33.56 & 34.74 &
78.12 & 29.54 & 30.15 \\

Milk &
59.92 & 
70.29 & 36.80 & 37.42 &
81.50 & 30.41 & 31.94 \\

Cocoa &
58.68 & 
69.04 & 35.40 & 37.25 &
78.54 & 29.14 & 31.94 \\

Chips &
61.57 & 
70.38 & 35.88 & 37.38 &
81.45 & 29.85 & 31.53 \\

Cappuccino &
60.74 & 
70.55 & 41.22 & 37.69 &
81.93 & 33.14 & 31.23 \\

\bottomrule
\end{tabular}}
\end{center}
\caption{
Error rates of compositional action recognition on \emph{H2O} $\mathcal{S}_1$ split. We report classification accuracy in $\%$ and hand error in $mm$.
}
\label{tbl:H2O_full_s1}
\end{table*}
\newcolumntype{C}{>{\centering\arraybackslash}X}
\begin{table*}[h]
\begin{center}
\resizebox{1\linewidth}{!}{
\begin{tabular}{l | c | c c c | c c c}
\toprule
  & MViTv$2$~\cite{li2022mvitv2} & \multicolumn{3}{c|}{HTT~\cite{wen2023hierarchical}} & \multicolumn{3}{c}{Ours} \\
 \midrule
 Unseen objects & \small{Top-$1$ ($\%$) $\uparrow$} & \small{Top-$1$ ($\%$) $\uparrow$} & \small{Hand (L) $\downarrow$} & \small{Hand (R) $\downarrow$}  & \small{Top-$1$ ($\%$) $\uparrow$} & \small{Hand (L) $\downarrow$} & \small{Hand (R) $\downarrow$}  \\

\midrule
Book + Cappucinno &
60.74 & 
61.09 & 44.61 & 41.20 &
70.87 & 33.45 & 32.52 \\

Espresso + Chips &
54.13 & 
62.60 & 40.18 &  42.17 &
71.45 & 31.14 & 34.89 \\

Lotion + Cocoa &
42.56 & 
58.58 & 40.28 & 39.94 &
69.52 & 32.58 & 33.42 \\

Spray + Milk &
48.35 & 
54.81 & 40.12 & 39.45 &
64.13 & 31.45 & 32.95 \\

Lotion + Spray &
54.55 & 
61.11 & 44.31 & 40.82 &
70.50 & 34.48 & 36.63 \\

Milk + Cocoa &
49.17 & 
59.45 & 42.21 & 39.66 &
69.95 & 32.12 & 32.52 \\

Cocoa + Chips &
51.65 & 
62.65 & 41.87 & 40.52 &
72.45 & 32.68 & 33.56 \\

Book + Spray &
57.44 & 
58.78 & 44.58 & 42.09 &
70.56 & 35.62 & 34.52 \\

\bottomrule
\end{tabular}}
\end{center}
\caption{
Error rates of compositional action recognition on \emph{H2O} $\mathcal{S}_2$ split. We report classification accuracy in $\%$ and hand error in $mm$.
}
\label{tbl:H2O_full_s2}
\end{table*}

\subsubsection{Superquadrics}\label{sup:sq}
\begin{figure*}[ht]
\centering
\includegraphics[width=1\linewidth]{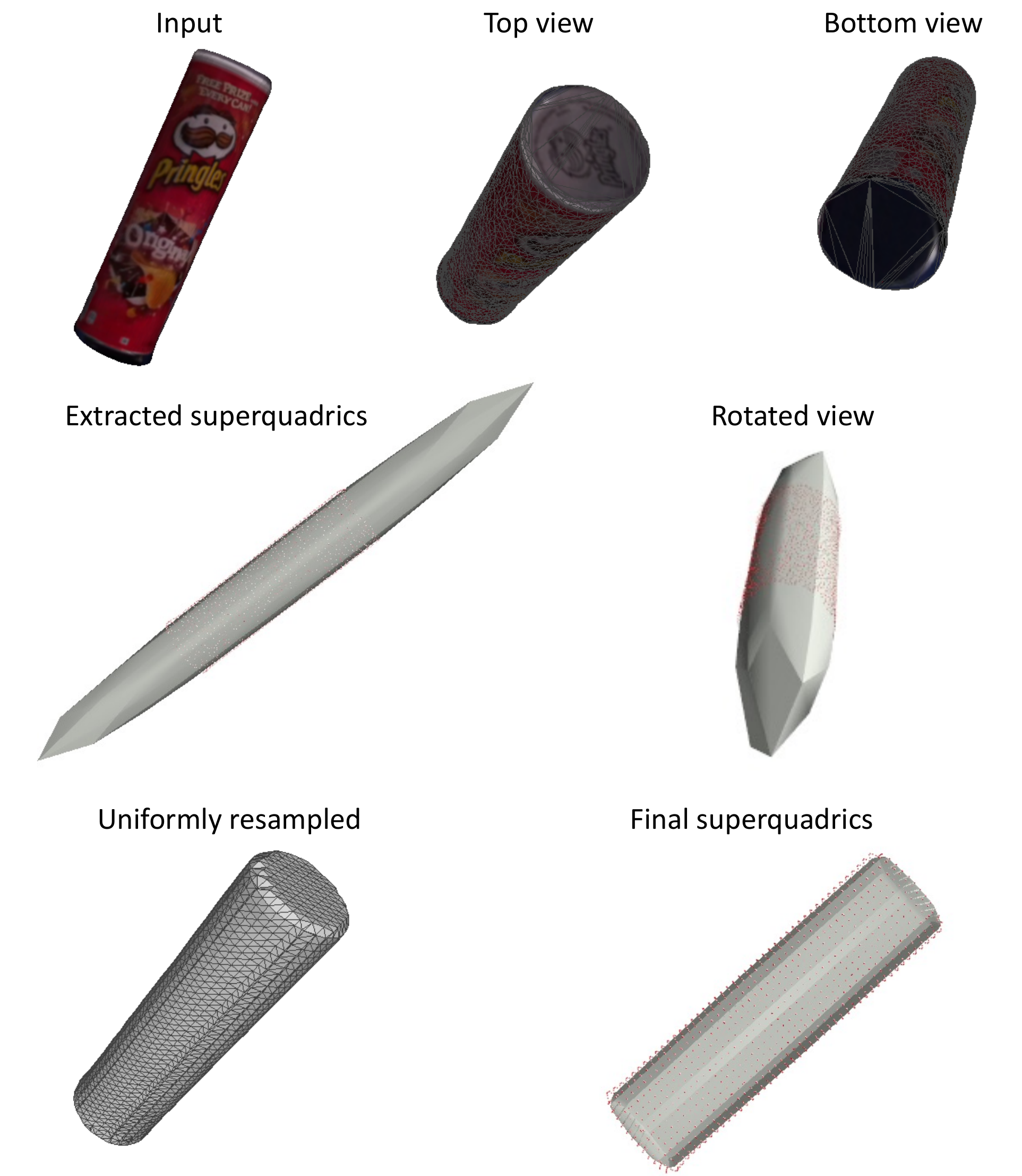}
\caption{Example failure case for superquadrics extraction. \textbf{First row:} We show input target object and the rotated views with wireframe. The visualisation of wireframe shows that the top and bottom of the input mesh is broken. \textbf{Second row:} We visualise the extracted superquadrics in grey and sampled point clouds in red. As the sampled point clouds are unable to capture the enclosed surface of the top and bottom, the estimated superquadric fails to represent the input object accurately. \textbf{Third row:} To fix this problem, we uniformly resample the input mesh such that it is watertight. We show that the resulting superquadric estimation can well-represent the target object.}
\label{fig:failure_SQ}
\end{figure*}

\paragraph{Data pre-processing.} We extract superquadrics from object models using the Expectation, Maximization and Switching (EMS) algorithm developed by~\citet{liu2022robust}. This is the first probabilistic method to recover superquadrics from noisy point clouds. We treat mesh vertices from object model as point clouds input to the algorithm and may further sample point clouds if the object template is not uniformly registered. The algorithm can be briefly described in 3 steps: 1) Infer the probability of a point being an outlier. 2) Given the current latent variables estimations, update the parameters of the superquadric. 3) Perform a global search for candidate parameters that encode similar superquadrics in terms of shape and pose and switch to the candidate that can increase the likelihood. We refer readers to~\citet{liu2022robust} for more details. In addition, we present an example failure case where a broken object model can heavily degrade the accuracy of superquadrics recovery in \fig{failure_SQ}. As illustrated in \fig{failure_SQ}, we show that the problem can be resolved by uniformly resampling the input object mesh.
\begin{figure}[h]
\centering
\includegraphics[width=0.9\linewidth]{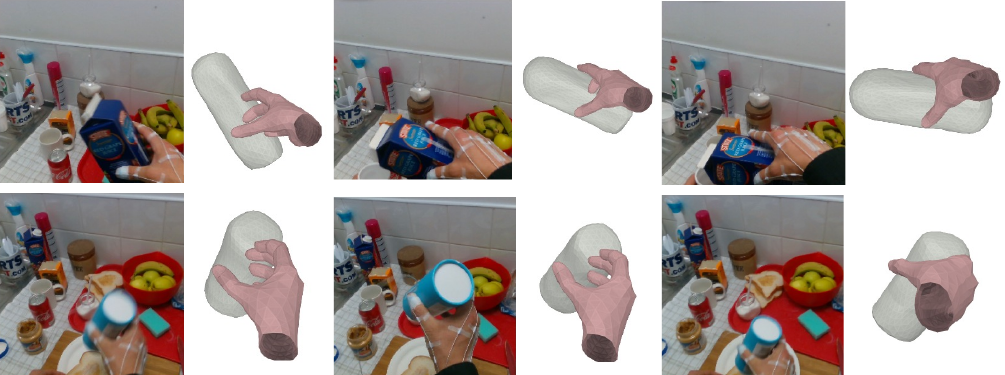}
\caption{Qualitative examples on \emph{FPHA}.}
\label{fig:fpha_sup}
\end{figure}                                 

\paragraph{Hand-object reconstruction.}
We additionally explore the performance of hand-object reconstruction of our method on \emph{ObMan} dataset. In Table~\ref{tbl:obman}, we report in 4 metrics: 
\begin{itemize}
    \item \emph{Hand error.} The mean end-point error ($mm$) over 21 joints.
    \item \emph{Object error.} By following~\cite{hasson2019learning,tse2022collaborative}, we measure the accuracy of object reconstruction by computing the Chamfer distance ($mm$) between points sampled on ground truth and predicted mesh.
    \item \emph{Penetration depth.} The maximum distances from hand mesh vertices to the object’s surface when in a collision.
    \item \emph{Intersection volume.} This is obtained by voxelising the hand and object using a voxel size of $0.5cm$.
    
\end{itemize}

\begin{table}[h]
\begin{center}
\resizebox{1\linewidth}{!}{
\begin{tabular}{l | ccc }\toprule
Method & \cite{hasson2019learning} & \cite{tse2022collaborative} & Ours\\ \hline

Hand error ($mm$) $\downarrow$ & 11.6 & 9.1 & \textbf{7.8} \\ 

Object error ($mm$) $\downarrow$ & 641.5 & 385.7 & \textbf{215.8}\\

Max. penetration ($mm$) $\downarrow$ & 9.5  & 7.4 & \textbf{5.1} \\ 

Intersection vol. ($cm^{3}$) $\downarrow$ & 12.3 & 9.3 & \textbf{4.5}\\ \bottomrule
\end{tabular}}
\end{center}
\caption{Quantitative comparison on \emph{ObMan}. Our proposed method with superquadrics outperforms existing state-of-the-arts.}
\label{tbl:obman}
\end{table}

\paragraph{Previously unseen objects.}
To verify the effectiveness of superquadrics on previously unseen objects for action recognition, we perform additional ablations on \emph{H2O} dataset. In Table~\ref{tbl:unseen_obj}, we experiment on the pose-based baseline HTT~\cite{wen2023hierarchical} and our proposed method by comparing the performance when with or without superquadrics. We estimate the object pose when without superquadrics, \ie w/o SQ. We observe that superquadrics effectively improve action recognition accuracy and recover better hand poses in both challenging settings.

\newcolumntype{C}{>{\centering\arraybackslash}X}
\begin{table}[h]
\begin{center}
\scalebox{1}{
\begin{tabular}{l | cc | cc }
\toprule
Method &  $\mathcal{S}_1 (\%) \uparrow$ & Hand $\downarrow$ &  $\mathcal{S}_2 (\%) \uparrow$ & Hand $\downarrow$\\
\midrule
HTT & 71.13 &  37.4 & 59.88 & 41.5 \\
HTT w. SQ & 73.25 & 35.1 & 62.45 & 39.4 \\
\midrule
Ours w/o SQ & 75.39 & 34.6 & 65.16 & 37.7 \\
Ours & 80.59 & 31.8 & 69.93 &  33.4 \\
\bottomrule
\end{tabular}%
}
\end{center}
\caption{Error rates of compositional action recognition on \emph{H2O}. We report the mean classification accuracy in $\%$ and hand error in $mm$ for both compositional split settings, $\mathcal{S}_1$ and $\mathcal{S}_2$. SQ refers to superquadrics and object pose is estimated when without superquadrics, \ie w/o SQ. We show that estimating superquadrics can consistently improve performance on the pose-based baseline HTT~\cite{wen2023hierarchical} and our proposed method.
}
\label{tbl:unseen_obj}
\end{table}

\subsection{Additional Examples} \label{sec:examples}

\paragraph{FPHA.} We show additional qualitative examples in \fig{fpha_sup}.

\begin{figure*}[h]
\centering
\includegraphics[width=1\linewidth]{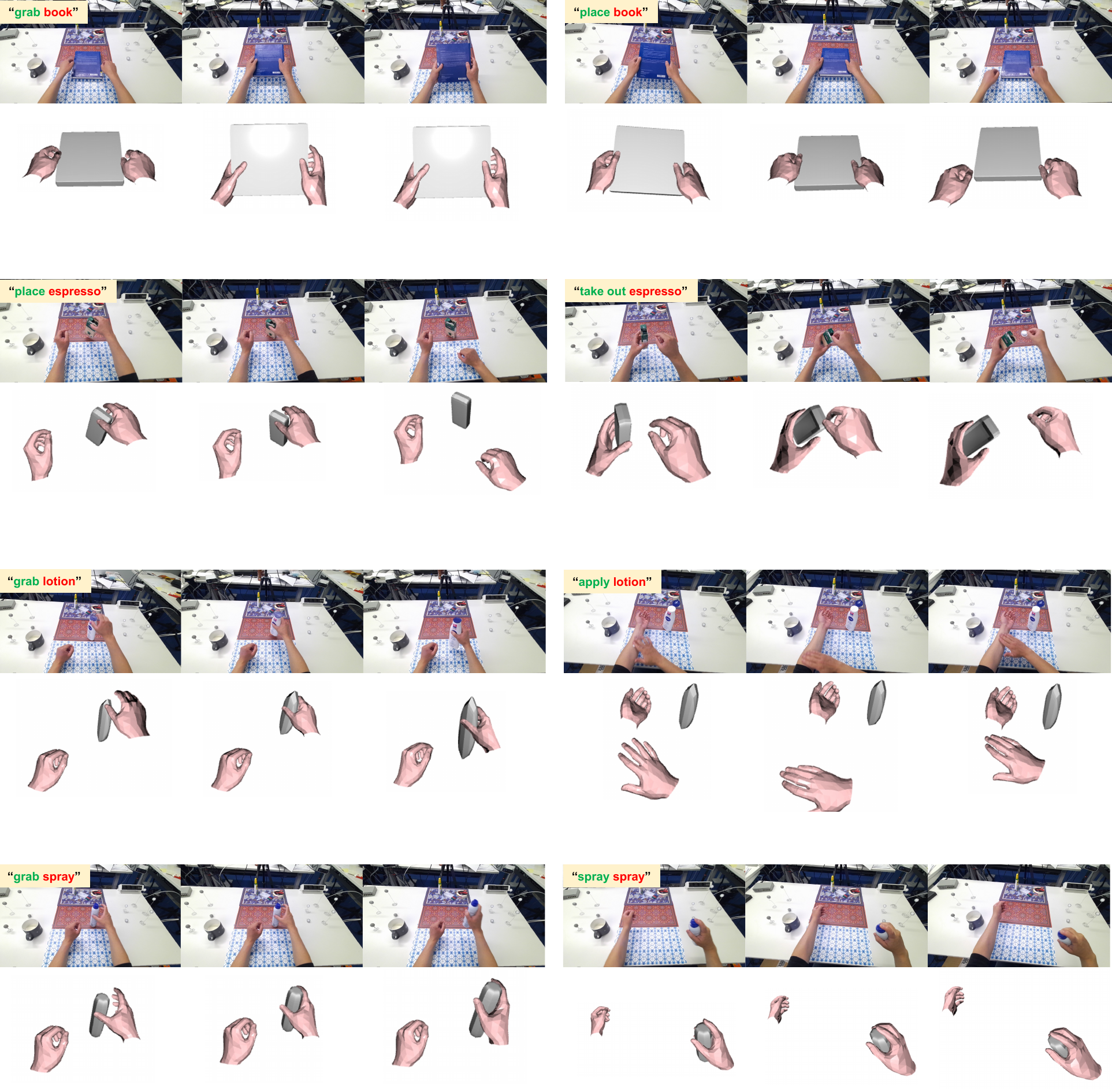}
\caption{Qualitative example on \emph{H2O}.}
\label{fig:h2o_1}
\end{figure*}

\begin{figure*}[h]
\centering
\includegraphics[width=1\linewidth]{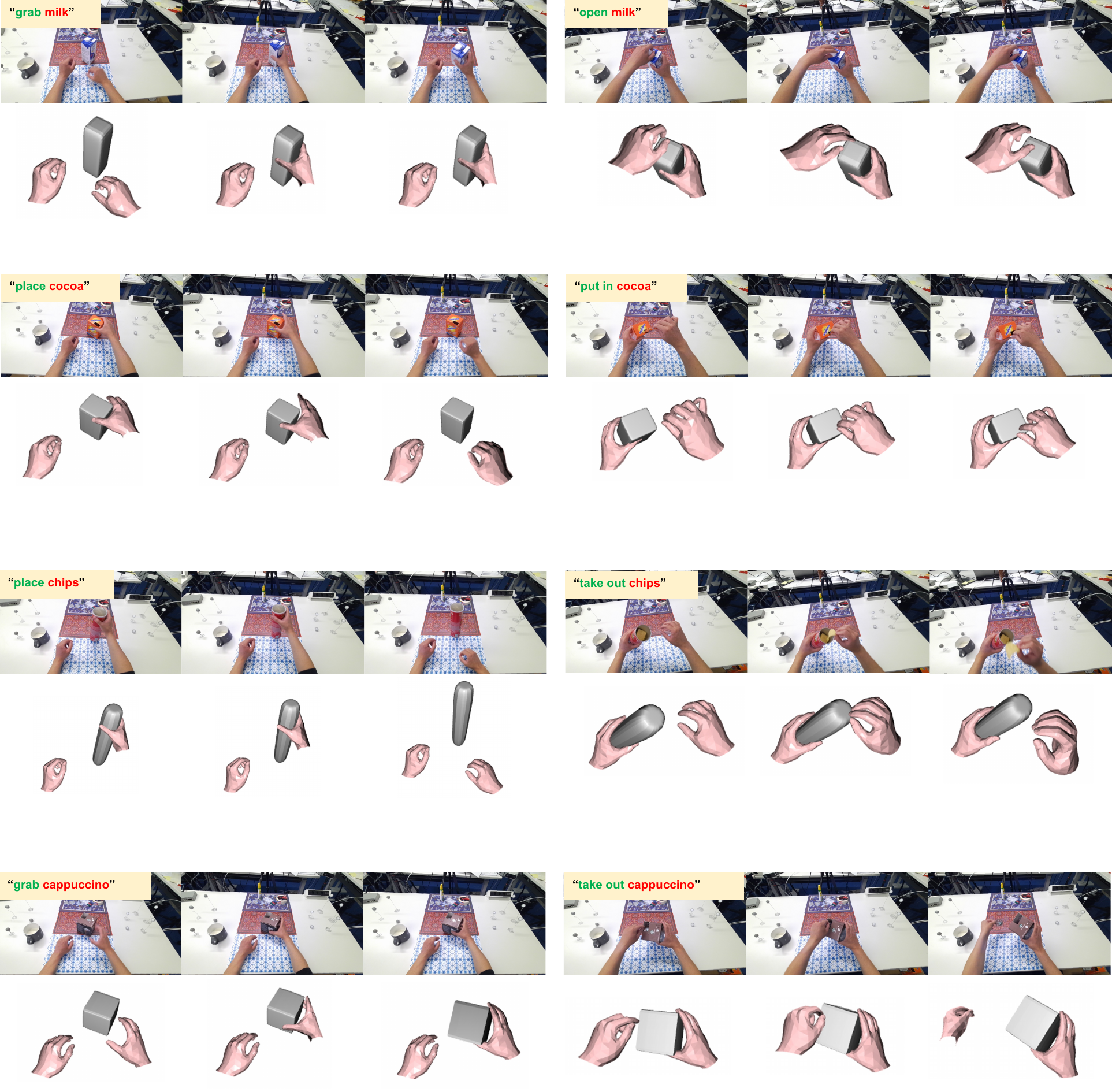}
\caption{Qualitative example on \emph{H2O}.}
\label{fig:h2o_2}
\end{figure*}

\paragraph{H2O.} We show additional qualitative examples in \fig{h2o_1} and \fig{h2o_2}. We include every objects in the dataset for both of the figures.


\subsection{Broader Impacts} \label{sup:broader}
This research on egocentric video understanding has both positive and negative societal impacts. On the positive side, it enables intelligent machines to comprehend video streams holistically, similar to how humans do. By correlating concepts and abstracting knowledge from different tasks, this research can enhance the learning of novel skills. However, there are challenges in executing multiple tasks with a single architecture and repurposing knowledge across tasks. Achieving human-like holistic reasoning remains a distant goal for artificial intelligence systems

\clearpage
\bibliography{aaai25}

\end{document}